\documentstyle[a4wide,11pt]{article}
\title{Conserving Fuel in Statistical Language Learning:\\
	Predicting Data Requirements
 \footnote{This paper has been
 accepted for publication at the Eigth Australian Joint Conference
 on Artificial Intelligence, Canberra, 1995.}
}

\author{Mark Lauer
\\ {\normalsize \it 65 Epping Road, North Ryde NSW 2113, Australia}
\\ {\normalsize Email: \tt t-markl@microsoft.com}
}

\date{\vspace{-0.3in}}

\begin{document}

\maketitle

\begin{abstract}

The paradigm for {\sc nlp} known as {\sc statistical language
learning} ({\sc sll}) has flourished in recent times, being seen
as a quick and easy way to get off the ground.
Research systems have been launched at many {\sc nlp}
problems including sense disambiguation (Yarowsky,~1992),
anaphora resolution (Dagan and Itai,~1990),
prepositional phrase attachment (Hindle and Rooth, 1993)
and lexical acquisition (Brent, 1993).  This has all been fueled by
the large text corpora which are
increasingly available (Marcus {\it et~al}.,~1993).
Since these systems learn to navigate language by consuming
text, they are critically dependent on the data that
drives them.

In this paper I address the practical concern of predicting
how much training data is sufficient for a given system.  First,
I briefly review earlier results and show how these can be
combined to bound the expected accuracy of a mode-based
learner as a function of the volume of training data.
I then develop a more accurate estimate of the expected
accuracy function under the assumption that inputs
are uniformly distributed.  Since this estimate is expensive
to compute, I also give a close but cheaply computable
approximation to it.  Finally, I report on a series of simulations
exploring the effects of inputs that are not uniformly
distributed.

\end{abstract}

\section{Background}

\subsection{Do We Need To Know?}

Even though text is becoming increasingly available, it is often
expensive, especially if it must be annotated.  Consider the
decisions facing the {\sc sll} technology consumer, that is,
the architect of a planned commercial {\sc nlp} system.
For each module which is to employ {\sc sll}, an appropriate technique
must be selected.  If different techniques require different
amounts of data to achieve a given accuracy,
the architect would like to know what these requirements are in advance
in order to make an informed choice.

Further, once the technique is chosen, she must decide how much
data to collect or purchase for training.  Because this data can be expensive,
foreknowledge of data requirements is highly valuable.
Thus, in order to make statistical {\sc nlp} technology practical,
a predictive theory of data requirements is needed.
Despite this need, very little attention has been paid to the
problem.\footnote{See de Haan (1992) for an investigation of sample
sizes for linguistic studies.}

\subsection{Foundations For A Theory}  \label{sec_foundations}
All the {\sc sll} systems mentioned above employ knowledge gained
from a corpus to make decisions.  Abstractly, this knowledge
can be represented as a mapping from observable features (inputs)
to decision outcomes (outputs).  Following Lauer (1995) I will
call each distinguished input a {\sc bin} and each possible output
a {\sc value}.  There is a probability distribution across the bins
representing how instances fall into bins.  Also, for each bin,
there is a probability distribution across the set of values
representing how instances in that bin take on values.
For the system to perform accurately, most (but not necessarily
all) of the instances falling in a particular bin must have the
same value.

In what follows I will make several assumptions: Training and test
data are drawn from the same distributions.  The set of possible
values is binary (examples include Hindle and Rooth, 1993 and
Lauer, 1994).  The probability of the most likely value in each
bin is constant.\footnote{Note that this does not require that the most
likely value be the same value in each bin; only that whatever
the most likely value is has a constant probability.}
Finally, I will only consider a simple learning algorithm: collect
the training instances falling into each bin and then select
the most frequent value for each.  This mode-based learner is
employed directly in the unigram tagger of Charniak (1993, p49)
and is at the heart of many systems.

\subsection{Optimal Accuracy}

There are two sources of error in statistical language
learners of the kind we are considering.  First, since the
values are not necessarily fully determined by the bins,
no matter what value the learner
assigns to a bin there will always be errors (the optimal
error rate).
Second, since training
data is limited, the learner may not have sufficient data
available to acquire accurate rules.
The combination of these sources of error results in some
degree of inaccuracy for the system.  We are interested in
estimating the accuracy for various volumes of training
data.  Since the optimal error rate is independent of the
amount of training data,  it will always exist no matter how
much data is used.
As the amount of training data increases we
expect the accuracy to get closer to this optimal.

Let $B$ be the set of bins, $V$ the set of values, $\Pr(b)$ the
probability that an instance falls into the bin $b$ and
$\Pr(v \mid b)$ the probability of the value $v$ given the
bin $b$.  If we denote the most likely value in each bin as
$v_b = \mbox{argmax}_{v \in V} \Pr(v \mid b)$, then the
expected value of the optimal accuracy is determined by
the likelihood of this value occurring in each bin.
\begin{equation}
\mbox{OA} = \sum_{b \in B} \Pr(b) \Pr(v_b \mid b)
\end{equation}

If we know the probability that an algorithm will learn the
value $v$ for the bin $b$ (denote this
$\Pr(\mbox{learn}(b)=v)$), then we can also calculate the
expected accuracy rate:
\begin{equation}
\mbox{EA} = \sum_{b \in B} \Pr(b) \sum_{v \in V}
\Pr(\mbox{learn}(b)=v) \Pr(v \mid b)
\end{equation}
In Lauer (1995) several results are shown concerning the
relationship of these two values.  I will summarise these
in section~\ref{sec_old_bounds} (see equations~(\ref{eqn_empty})
and (\ref{eqn_ea2oa})).

\section{Existing Work}

\subsection{Empty Bins and Non-empty Bins}  \label{sec_old_bounds}

The most severe result of insufficient training data is that
some bins can go without any training instances.  Since the
learner has no indications about likely values for the bin it
will be forced to guess.  To estimate how often this will
occur, consider the way in which $m$ training instances
would fall into the bins.  For each bin, the probability
that no training instances fall into it is:
\begin{displaymath}
\Pr(\mbox{count}(b) = 0) = (1-\Pr(b))^m
\end{displaymath}
I will call such bins {\sc empty bins}.

In Lauer (1995) it is shown that for any bin $b$:
\begin{equation}
\Pr(\mbox{count}(b) = 0) < e^{-m/{\mid B \mid}}
		\label{eqn_empty}
\end{equation}

Lauer (1995) also bounds the expected
accuracy of the mode-based learner when all bins are
guaranteed to have at least one training instance.  When
this is the case, it is shown that the expected error rate
is always no worse than twice the optimal error rate.
\begin{equation}
\mbox{EA} \ge (1-2(1-\mbox{OA}))
		\label{eqn_ea2oa}
\end{equation}

This is quite a useful result, since we expect the optimal
accuracy to be fairly high.  If the optimal
predictions are 90\% accurate, then a mode-based learner
will be at least 80\% accurate after learning on just
one instance per bin.

\subsection{Overall Expected Accuracy}

Unfortunately, we cannot normally guarantee that no bins
will be empty, since the corpus is typically a random
sample.  However, we can combine equations~(\ref{eqn_empty})
and~(\ref{eqn_ea2oa}) to arrive at a bound for the overall expected
accuracy after training on a random sample.
Over non-empty bins, we know that the error rate is no worse than
twice the optimal error rate for those bins.
Since we have assumed that $\Pr(v_b \mid b)$ is constant
(call this $p$), we can infer
that the optimal accuracy for the non-empty bins is
the same as the optimal accuracy on all bins.  Thus:
\begin{eqnarray}
\mbox{EA} & = & \Pr(\mbox{non-empty}) \mbox{EA}(\mbox{non-empty})
		+ \Pr(\mbox{empty}) \mbox{EA}(\mbox{empty}) \nonumber \\
	& \ge & (1-e^{-m/{\mid B \mid}}) \mbox{EA}(\mbox{non-empty})
		+ (e^{-m/{\mid B \mid}}) \mbox{EA}(\mbox{empty}) \nonumber \\
	& \ge & (1-e^{-m/{\mid B \mid}}) (1-2(1-\mbox{OA}))
		+ \frac{1}{2}e^{-m/{\mid B \mid}} \nonumber \\
	& = & (1-e^{-m/{\mid B \mid}}) (2p-1)
		+ \frac{1}{2}e^{-m/{\mid B \mid}} \label{eqn_old_bound}
\end{eqnarray}
The second step follows from the fact that
$\mbox{EA}(\mbox{non-empty}) \ge \mbox{EA}(\mbox{empty})$ and
equation~(\ref{eqn_empty}).  The third step follows from
equation~(\ref{eqn_ea2oa}).

\section{Theory}  \label{sec_theory}

\subsection{Estimating Expected Accuracy}

Given the assumptions in section~\ref{sec_foundations}, we
can arrive at a better estimate of the expected accuracy
when the distribution of bins is uniform (that is,
$\Pr(b) = \frac{1}{\mid B \mid}$).
Let the total number of training instances in a bin $b$ be $n$
and the number of these instances with value $v$ be
$\mbox{count}(v,b)$:
\begin{displaymath}
\Pr(\mbox{count}(v,b) > n/2) = \sum_{i=\lceil {n+1}/2
\rceil}^{n} {n \choose i} {\Pr(v \mid b)}^i (1-\Pr(v \mid
b))^{n-i}
\end{displaymath}
If $n$ is even, we must also add an additional term of
$1/2 {n \choose n/2} {\Pr(v \mid b)}^{n/2} (1-\Pr(v \mid
b))^{n/2}$.  This is because when there are equal numbers
of both values in the bin, a random guess yields an
expected accuracy of 50\%.  In the arguments below, I will
treat all values of $n$ as odd in order to simplify.  The
reader may check for herself that the results hold generally
when the above extra term is included.

Using the fact that $V$ is binary, the total expected accuracy
for test instances in bin $b$
when it contains $n$ training instances is:
\begin{displaymath}
\Pr(v = \mbox{argmax}_{v' \in V} \mbox{count}(v',b)) =
        \sum_{v \in V} \Pr(v \mid b)
	    \sum_{i=(n+1)/2}^{n}
		  {n \choose i} {\Pr(v \mid b)}^i (1-\Pr(v \mid b))^{n-i}
\end{displaymath}
By summing over all possible numbers of training
instances in a bin, we can arrive at an expression for the
expected accuracy across all bins as follows:
\begin{displaymath}
\mbox{EA} =
    \sum_{b \in B} \Pr(b)
        \sum_{n=0}^m \mbox{binomial}(n;m, \Pr(b))
            \sum_{v \in V} \Pr(v \mid b)
                \sum_{i=(n+1)/2}^{n}
		    \mbox{binomial}(i;n, \Pr(v \mid b))
\end{displaymath}
where $\mbox{binomial}(j;k,p) = {k \choose j} p^j (1-
p)^{k-j}$.

To simplify this I have defined a function as follows:
\begin{displaymath}
G(m,r,p) = \sum_{n=0}^m \mbox{binomial}(n;m,r)
                      \sum_{i=(n+1)/2}^{n}
\mbox{binomial}(i;n,p)
\end{displaymath}
A result which may be easily obtained by expansion is:
\begin{equation}
G(m,r,1-p) = 1 - G(m,r,p)   \label{eqn_Gcompl}
\end{equation}
Using the assumptions in section~\ref{sec_foundations} and the
uniform bin probabilities we can now proceed to simplify:
\begin{eqnarray}
\mbox{EA} & = &
    \sum_{b \in B} \frac{1}{\mid B \mid}
        \sum_{n=0}^m \mbox{binomial}(n;m, \frac{1}{\mid B \mid})
            \sum_{v \in V} \Pr(v \mid b)
                \sum_{i=(n+1)/2}^{n}
		    \mbox{binomial}(i;n, \Pr(v \mid b))  \nonumber \\
& = &   \sum_{b \in B} \frac{1}{\mid B \mid}
            \sum_{v \in V} \Pr(v \mid b)
                \sum_{n=0}^m \mbox{binomial}(n;m, \frac{1}{\mid B \mid})
                    \sum_{i=(n+1)/2}^{n} \mbox{binomial}(i;n, \Pr(v \mid b))
			\nonumber \\
& = &   \sum_{b \in B} \frac{1}{\mid B \mid}
            \sum_{v \in V} \Pr(v \mid b)
                G(m, \frac{1}{\mid B \mid}, \Pr(v \mid b))
			\nonumber \\
& = &   \sum_{b \in B} \frac{1}{\mid B \mid}
        (p G(m, \frac{1}{\mid B \mid}, p)
        + (1-p) G(m, \frac{1}{\mid B \mid}, 1-p)) \nonumber \\
& = &   (1-p + (2p-1) G(m, \frac{1}{\mid B \mid}, p))
\label{eqn_ea_bound}
\end{eqnarray}
The last step uses equation~(\ref{eqn_Gcompl}) and
$\sum_{b \in B} \frac{1}{\mid B \mid} = 1$.

\subsection{A Computable Bound for G}  \label{sec_computable}

The main difficulty with the function $G$ is the
appearance of ${m \choose n}$.
Most corpus-based language learners
use large corpora, so we expect the number of training
instances, $m$, to be very large.
So we need a
more easily computable version of $G$.  The following argument
leads to a fairly tight lower bound to $G$ for suitably chosen
values of $k_j$ (see below):
\begin{eqnarray*}
G(m,r,p)
	& = & \sum_{n=0}^m \sum_{i=(n+1)/2}^{n}
                      \mbox{binomial}(n;m,r)
\mbox{binomial}(i;n,p)   \\
	& = & \sum_{j=0}^{(m-1)/2} \sum_{n=2j+1}^m
                      \mbox{binomial}(n;m,r)
\mbox{binomial}(n-j;n,p)   \\
	& = & \sum_{j=0}^{(m-1)/2} \sum_{n=2j+1}^m
                      {m \choose n} r^n (1-r)^{m-n} {n \choose j}
p^{n-j} (1-p)^j   \\
	& = & \sum_{j=0}^{(m-1)/2} (1-r)^m (\frac{1-
p}{p})^j
                      \sum_{n=2j+1}^m \frac{m!}{(m-n)!} (1-
r)^{-n}
                          \frac{p^n}{n!} {n \choose j} r^n   \\
	& \ge & \sum_{j=0}^{(m-1)/2} (1-r)^m (\frac{1-
p}{p})^j
                      \sum_{n=2j+1}^{k_j} \frac{m!}{(m-n)!} (1-
r)^{-n}
                          \frac{p^n}{n!} {n \choose j} r^n
\end{eqnarray*}
The first step rearranges the order of addition.  The final
step introduces a series of variables which limit the
number of terms in the inner sum.  The inequality holds for
all $k_j \le m$.  Notice that the $k_j$ may vary for each
term of the outer sum.  Since $n \le k_j \le m$ we can use
the following relation:
\begin{equation}
\frac{m!}{(m-n)!} \ge (m-k_j)^n
\label{eqn_factorial_approx}
 \end{equation}
Letting $x_j = r p \frac{(m-k_j)}{(1-r)}$ we can simplify as
follows:
\begin{eqnarray*}
G(m,r,p) & \ge & \sum_{j=0}^{(m-1)/2} (1-r)^m (\frac{1-
p}{p})^j
                      \sum_{n=2j+1}^{k_j} {n \choose j}
\frac{m!}{(m-n)!}
                          \frac{(1-r)^{-n} r^n p^n}{n!}   \\
              & \ge & \sum_{j=0}^{(m-1)/2} (1-r)^m (\frac{1-
p}{p})^j
                      \sum_{n=2j+1}^{k_j} {n \choose j} (m-
k_j)^n
                          \frac{(1-r)^{-n} r^n p^n}{n!}   \\
               & = & \sum_{j=0}^{(m-1)/2} (1-r)^m (\frac{1-
p}{p})^j
                      \sum_{n=2j+1}^{k_j} {n \choose j}
\frac{x_j^n}{n!}  \\
               & \ge & \sum_{j=0}^{g} (1-r)^m (\frac{1-p}{p})^j
                      \sum_{n=2j+1}^{k_j} {n \choose j}
\frac{x_j^n}{n!}
\end{eqnarray*}
The last step introduces $g$ and holds for all $g \le (m-
1)/2$.  This is because in practice only the first few terms
of the outer sum are significant.  Thus for suitably chosen
$g, k_j$ this is a cheaply computable lower bound for
$G$.  A program to compute this to a high degree of
accuracy has been implemented.

\section{Experiment}

\subsection{Skewed Bins}  \label{sec_skewed}

The assumption that bin probabilities are uniform is problematic.
When bins are uniformly probable, the expected number of
training instances in the same bin as a random test instance is
$\frac{m}{\mid B \mid}$ ($ = \sum_{b \in B} \Pr(b) \sum_{n=0}^{m}
	n \Pr(n \mbox{ training items fall into } b)$).
But most distributions in language are highly skewed.
Zipf's law states that word types are distributed logarithmically
(the $n$th most frequent word has probability proportional
to $\frac{1}{n}$).  When this is true the expected number of
training instances in the same bin as a random test instance is
approximately $\frac{1.6 m}{\log{(0.56 \mid B \mid)}^2}$
($ \gg \frac{m}{\mid B \mid}$).  Thus we can expect much more
information to be available about typical test cases.

\subsection{Simulations}

Since the mathematics in section~\ref{sec_theory} cannot
easily be generalised to different distributions,
I have conducted several simulations in order to verify
the mathematical results above and to explore the effect
of using a skewed distribution of bins.

These simulations use a fixed number of bins (10,000), allocating
$m$ training instances to the bins according to either
a uniform or logarithmic distribution.  It then measures the
correctness of the mode-based learner on 1000 randomly generated
test instances to arrive at an observed correctness
rate.\footnote{The results were generated using an optimal
value probability of $p = 0.9$ (thus the optimal accuracy rate is 90\%).
Simulations with other values of $p$ did not differ qualitatively.}

This process (training and testing) is repeated 30 times for each run,
with the mean being recorded as the observed accuracy.
The standard deviation is used to estimate a 5\% t-score confidence interval.

\begin{figure}
\centering
\setlength{\unitlength}{0.240900pt}
\ifx\plotpoint\undefined\newsavebox{\plotpoint}\fi
\sbox{\plotpoint}{\rule[-0.200pt]{0.400pt}{0.400pt}}%
\begin{picture}(1500,900)(0,0)
\font\gnuplot=cmr10 at 10pt
\gnuplot
\sbox{\plotpoint}{\rule[-0.200pt]{0.400pt}{0.400pt}}%
\put(220.0,113.0){\rule[-0.200pt]{0.400pt}{184.048pt}}
\put(220.0,113.0){\rule[-0.200pt]{4.818pt}{0.400pt}}
\put(198,113){\makebox(0,0)[r]{50}}
\put(1416.0,113.0){\rule[-0.200pt]{4.818pt}{0.400pt}}
\put(220.0,189.0){\rule[-0.200pt]{4.818pt}{0.400pt}}
\put(198,189){\makebox(0,0)[r]{55}}
\put(1416.0,189.0){\rule[-0.200pt]{4.818pt}{0.400pt}}
\put(220.0,266.0){\rule[-0.200pt]{4.818pt}{0.400pt}}
\put(198,266){\makebox(0,0)[r]{60}}
\put(1416.0,266.0){\rule[-0.200pt]{4.818pt}{0.400pt}}
\put(220.0,342.0){\rule[-0.200pt]{4.818pt}{0.400pt}}
\put(198,342){\makebox(0,0)[r]{65}}
\put(1416.0,342.0){\rule[-0.200pt]{4.818pt}{0.400pt}}
\put(220.0,419.0){\rule[-0.200pt]{4.818pt}{0.400pt}}
\put(198,419){\makebox(0,0)[r]{70}}
\put(1416.0,419.0){\rule[-0.200pt]{4.818pt}{0.400pt}}
\put(220.0,495.0){\rule[-0.200pt]{4.818pt}{0.400pt}}
\put(198,495){\makebox(0,0)[r]{75}}
\put(1416.0,495.0){\rule[-0.200pt]{4.818pt}{0.400pt}}
\put(220.0,571.0){\rule[-0.200pt]{4.818pt}{0.400pt}}
\put(198,571){\makebox(0,0)[r]{80}}
\put(1416.0,571.0){\rule[-0.200pt]{4.818pt}{0.400pt}}
\put(220.0,648.0){\rule[-0.200pt]{4.818pt}{0.400pt}}
\put(198,648){\makebox(0,0)[r]{85}}
\put(1416.0,648.0){\rule[-0.200pt]{4.818pt}{0.400pt}}
\put(220.0,724.0){\rule[-0.200pt]{4.818pt}{0.400pt}}
\put(198,724){\makebox(0,0)[r]{90}}
\put(1416.0,724.0){\rule[-0.200pt]{4.818pt}{0.400pt}}
\put(220.0,801.0){\rule[-0.200pt]{4.818pt}{0.400pt}}
\put(198,801){\makebox(0,0)[r]{95}}
\put(1416.0,801.0){\rule[-0.200pt]{4.818pt}{0.400pt}}
\put(220.0,877.0){\rule[-0.200pt]{4.818pt}{0.400pt}}
\put(198,877){\makebox(0,0)[r]{100}}
\put(1416.0,877.0){\rule[-0.200pt]{4.818pt}{0.400pt}}
\put(220.0,113.0){\rule[-0.200pt]{0.400pt}{4.818pt}}
\put(220,68){\makebox(0,0){0}}
\put(220.0,857.0){\rule[-0.200pt]{0.400pt}{4.818pt}}
\put(394.0,113.0){\rule[-0.200pt]{0.400pt}{4.818pt}}
\put(394,68){\makebox(0,0){10000}}
\put(394.0,857.0){\rule[-0.200pt]{0.400pt}{4.818pt}}
\put(567.0,113.0){\rule[-0.200pt]{0.400pt}{4.818pt}}
\put(567,68){\makebox(0,0){20000}}
\put(567.0,857.0){\rule[-0.200pt]{0.400pt}{4.818pt}}
\put(741.0,113.0){\rule[-0.200pt]{0.400pt}{4.818pt}}
\put(741,68){\makebox(0,0){30000}}
\put(741.0,857.0){\rule[-0.200pt]{0.400pt}{4.818pt}}
\put(915.0,113.0){\rule[-0.200pt]{0.400pt}{4.818pt}}
\put(915,68){\makebox(0,0){40000}}
\put(915.0,857.0){\rule[-0.200pt]{0.400pt}{4.818pt}}
\put(1089.0,113.0){\rule[-0.200pt]{0.400pt}{4.818pt}}
\put(1089,68){\makebox(0,0){50000}}
\put(1089.0,857.0){\rule[-0.200pt]{0.400pt}{4.818pt}}
\put(1262.0,113.0){\rule[-0.200pt]{0.400pt}{4.818pt}}
\put(1262,68){\makebox(0,0){60000}}
\put(1262.0,857.0){\rule[-0.200pt]{0.400pt}{4.818pt}}
\put(1436.0,113.0){\rule[-0.200pt]{0.400pt}{4.818pt}}
\put(1436,68){\makebox(0,0){70000}}
\put(1436.0,857.0){\rule[-0.200pt]{0.400pt}{4.818pt}}
\put(220.0,113.0){\rule[-0.200pt]{292.934pt}{0.400pt}}
\put(1436.0,113.0){\rule[-0.200pt]{0.400pt}{184.048pt}}
\put(220.0,877.0){\rule[-0.200pt]{292.934pt}{0.400pt}}
\put(45,495){\makebox(0,0){\shortstack{Accuracy\\(\%)}}}
\put(828,-22){\makebox(0,0){Training set size ($m$)}}
\put(220.0,113.0){\rule[-0.200pt]{0.400pt}{184.048pt}}
\put(1262,342){\makebox(0,0)[r]{Optimal}}
\put(1284.0,342.0){\rule[-0.200pt]{15.899pt}{0.400pt}}
\put(220,724){\usebox{\plotpoint}}
\put(220.0,724.0){\rule[-0.200pt]{292.934pt}{0.400pt}}
\put(1262,297){\makebox(0,0)[r]{Logarithmic}}
\put(1284.0,297.0){\rule[-0.200pt]{15.899pt}{0.400pt}}
\put(220,113){\usebox{\plotpoint}}
\multiput(220.58,113.00)(0.496,8.195){41}{\rule{0.120pt}{6.555pt}}
\multiput(219.17,113.00)(22.000,341.396){2}{\rule{0.400pt}{3.277pt}}
\multiput(242.58,468.00)(0.496,1.152){39}{\rule{0.119pt}{1.014pt}}
\multiput(241.17,468.00)(21.000,45.895){2}{\rule{0.400pt}{0.507pt}}
\multiput(263.00,516.58)(0.611,0.498){69}{\rule{0.589pt}{0.120pt}}
\multiput(263.00,515.17)(42.778,36.000){2}{\rule{0.294pt}{0.400pt}}
\multiput(307.00,552.58)(0.855,0.498){99}{\rule{0.782pt}{0.120pt}}
\multiput(307.00,551.17)(85.376,51.000){2}{\rule{0.391pt}{0.400pt}}
\multiput(394.00,603.58)(2.565,0.498){65}{\rule{2.135pt}{0.120pt}}
\multiput(394.00,602.17)(168.568,34.000){2}{\rule{1.068pt}{0.400pt}}
\multiput(567.00,637.58)(4.923,0.495){33}{\rule{3.967pt}{0.119pt}}
\multiput(567.00,636.17)(165.767,18.000){2}{\rule{1.983pt}{0.400pt}}
\multiput(741.00,655.58)(4.206,0.496){39}{\rule{3.414pt}{0.119pt}}
\multiput(741.00,654.17)(166.913,21.000){2}{\rule{1.707pt}{0.400pt}}
\multiput(915.00,676.58)(9.043,0.491){17}{\rule{7.060pt}{0.118pt}}
\multiput(915.00,675.17)(159.347,10.000){2}{\rule{3.530pt}{0.400pt}}
\multiput(1089.00,686.58)(8.991,0.491){17}{\rule{7.020pt}{0.118pt}}
\multiput(1089.00,685.17)(158.430,10.000){2}{\rule{3.510pt}{0.400pt}}
\multiput(1262.00,696.59)(10.106,0.489){15}{\rule{7.833pt}{0.118pt}}
\multiput(1262.00,695.17)(157.742,9.000){2}{\rule{3.917pt}{0.400pt}}
\put(1306,297){\circle*{12}}
\put(220,113){\circle*{12}}
\put(242,468){\circle*{12}}
\put(263,516){\circle*{12}}
\put(307,552){\circle*{12}}
\put(394,603){\circle*{12}}
\put(567,637){\circle*{12}}
\put(741,655){\circle*{12}}
\put(915,676){\circle*{12}}
\put(1089,686){\circle*{12}}
\put(1262,696){\circle*{12}}
\put(1436,705){\circle*{12}}
\put(220,113){\circle*{12}}
\put(242,468){\circle*{12}}
\put(263,516){\circle*{12}}
\put(307,552){\circle*{12}}
\put(394,603){\circle*{12}}
\put(567,637){\circle*{12}}
\put(741,655){\circle*{12}}
\put(915,676){\circle*{12}}
\put(1089,686){\circle*{12}}
\put(1262,696){\circle*{12}}
\put(1436,705){\circle*{12}}
\put(220,113){\usebox{\plotpoint}}
\put(210.0,113.0){\rule[-0.200pt]{4.818pt}{0.400pt}}
\put(210.0,113.0){\rule[-0.200pt]{4.818pt}{0.400pt}}
\put(242.0,461.0){\rule[-0.200pt]{0.400pt}{3.373pt}}
\put(232.0,461.0){\rule[-0.200pt]{4.818pt}{0.400pt}}
\put(232.0,475.0){\rule[-0.200pt]{4.818pt}{0.400pt}}
\put(263.0,509.0){\rule[-0.200pt]{0.400pt}{3.613pt}}
\put(253.0,509.0){\rule[-0.200pt]{4.818pt}{0.400pt}}
\put(253.0,524.0){\rule[-0.200pt]{4.818pt}{0.400pt}}
\put(307.0,546.0){\rule[-0.200pt]{0.400pt}{2.891pt}}
\put(297.0,546.0){\rule[-0.200pt]{4.818pt}{0.400pt}}
\put(297.0,558.0){\rule[-0.200pt]{4.818pt}{0.400pt}}
\put(394.0,595.0){\rule[-0.200pt]{0.400pt}{3.613pt}}
\put(384.0,595.0){\rule[-0.200pt]{4.818pt}{0.400pt}}
\put(384.0,610.0){\rule[-0.200pt]{4.818pt}{0.400pt}}
\put(567.0,632.0){\rule[-0.200pt]{0.400pt}{2.409pt}}
\put(557.0,632.0){\rule[-0.200pt]{4.818pt}{0.400pt}}
\put(557.0,642.0){\rule[-0.200pt]{4.818pt}{0.400pt}}
\put(741.0,650.0){\rule[-0.200pt]{0.400pt}{2.409pt}}
\put(731.0,650.0){\rule[-0.200pt]{4.818pt}{0.400pt}}
\put(731.0,660.0){\rule[-0.200pt]{4.818pt}{0.400pt}}
\put(915.0,668.0){\rule[-0.200pt]{0.400pt}{3.613pt}}
\put(905.0,668.0){\rule[-0.200pt]{4.818pt}{0.400pt}}
\put(905.0,683.0){\rule[-0.200pt]{4.818pt}{0.400pt}}
\put(1089.0,680.0){\rule[-0.200pt]{0.400pt}{2.891pt}}
\put(1079.0,680.0){\rule[-0.200pt]{4.818pt}{0.400pt}}
\put(1079.0,692.0){\rule[-0.200pt]{4.818pt}{0.400pt}}
\put(1262.0,690.0){\rule[-0.200pt]{0.400pt}{3.132pt}}
\put(1252.0,690.0){\rule[-0.200pt]{4.818pt}{0.400pt}}
\put(1252.0,703.0){\rule[-0.200pt]{4.818pt}{0.400pt}}
\put(1436.0,700.0){\rule[-0.200pt]{0.400pt}{2.650pt}}
\put(1426.0,700.0){\rule[-0.200pt]{4.818pt}{0.400pt}}
\put(1426.0,711.0){\rule[-0.200pt]{4.818pt}{0.400pt}}
\put(1262,252){\makebox(0,0)[r]{Uniform}}
\put(1284.0,252.0){\rule[-0.200pt]{15.899pt}{0.400pt}}
\put(220,113){\usebox{\plotpoint}}
\multiput(220.58,113.00)(0.496,1.446){41}{\rule{0.120pt}{1.245pt}}
\multiput(219.17,113.00)(22.000,60.415){2}{\rule{0.400pt}{0.623pt}}
\multiput(242.58,176.00)(0.496,1.080){39}{\rule{0.119pt}{0.957pt}}
\multiput(241.17,176.00)(21.000,43.013){2}{\rule{0.400pt}{0.479pt}}
\multiput(263.58,221.00)(0.498,1.003){85}{\rule{0.120pt}{0.900pt}}
\multiput(262.17,221.00)(44.000,86.132){2}{\rule{0.400pt}{0.450pt}}
\multiput(307.58,309.00)(0.499,0.690){171}{\rule{0.120pt}{0.652pt}}
\multiput(306.17,309.00)(87.000,118.647){2}{\rule{0.400pt}{0.326pt}}
\multiput(394.00,429.58)(0.704,0.499){243}{\rule{0.663pt}{0.120pt}}
\multiput(394.00,428.17)(171.625,123.000){2}{\rule{0.331pt}{0.400pt}}
\multiput(567.00,552.58)(1.090,0.499){157}{\rule{0.970pt}{0.120pt}}
\multiput(567.00,551.17)(171.987,80.000){2}{\rule{0.485pt}{0.400pt}}
\multiput(741.00,632.58)(2.579,0.498){65}{\rule{2.147pt}{0.120pt}}
\multiput(741.00,631.17)(169.544,34.000){2}{\rule{1.074pt}{0.400pt}}
\multiput(915.00,666.58)(3.522,0.497){47}{\rule{2.884pt}{0.120pt}}
\multiput(915.00,665.17)(168.014,25.000){2}{\rule{1.442pt}{0.400pt}}
\multiput(1089.00,691.58)(6.840,0.493){23}{\rule{5.423pt}{0.119pt}}
\multiput(1089.00,690.17)(161.744,13.000){2}{\rule{2.712pt}{0.400pt}}
\multiput(1262.00,704.59)(13.231,0.485){11}{\rule{10.043pt}{0.117pt}}
\multiput(1262.00,703.17)(153.156,7.000){2}{\rule{5.021pt}{0.400pt}}
\put(1306,252){\circle{12}}
\put(220,113){\circle{12}}
\put(242,176){\circle{12}}
\put(263,221){\circle{12}}
\put(307,309){\circle{12}}
\put(394,429){\circle{12}}
\put(567,552){\circle{12}}
\put(741,632){\circle{12}}
\put(915,666){\circle{12}}
\put(1089,691){\circle{12}}
\put(1262,704){\circle{12}}
\put(1436,711){\circle{12}}
\put(220,113){\circle{12}}
\put(242,176){\circle{12}}
\put(263,221){\circle{12}}
\put(307,309){\circle{12}}
\put(394,429){\circle{12}}
\put(567,552){\circle{12}}
\put(741,632){\circle{12}}
\put(915,666){\circle{12}}
\put(1089,691){\circle{12}}
\put(1262,704){\circle{12}}
\put(1436,711){\circle{12}}
\put(220,113){\usebox{\plotpoint}}
\put(210.0,113.0){\rule[-0.200pt]{4.818pt}{0.400pt}}
\put(210.0,113.0){\rule[-0.200pt]{4.818pt}{0.400pt}}
\put(242.0,169.0){\rule[-0.200pt]{0.400pt}{3.373pt}}
\put(232.0,169.0){\rule[-0.200pt]{4.818pt}{0.400pt}}
\put(232.0,183.0){\rule[-0.200pt]{4.818pt}{0.400pt}}
\put(263.0,211.0){\rule[-0.200pt]{0.400pt}{4.818pt}}
\put(253.0,211.0){\rule[-0.200pt]{4.818pt}{0.400pt}}
\put(253.0,231.0){\rule[-0.200pt]{4.818pt}{0.400pt}}
\put(307.0,299.0){\rule[-0.200pt]{0.400pt}{4.818pt}}
\put(297.0,299.0){\rule[-0.200pt]{4.818pt}{0.400pt}}
\put(297.0,319.0){\rule[-0.200pt]{4.818pt}{0.400pt}}
\put(394.0,420.0){\rule[-0.200pt]{0.400pt}{4.095pt}}
\put(384.0,420.0){\rule[-0.200pt]{4.818pt}{0.400pt}}
\put(384.0,437.0){\rule[-0.200pt]{4.818pt}{0.400pt}}
\put(567.0,545.0){\rule[-0.200pt]{0.400pt}{3.373pt}}
\put(557.0,545.0){\rule[-0.200pt]{4.818pt}{0.400pt}}
\put(557.0,559.0){\rule[-0.200pt]{4.818pt}{0.400pt}}
\put(741.0,624.0){\rule[-0.200pt]{0.400pt}{3.613pt}}
\put(731.0,624.0){\rule[-0.200pt]{4.818pt}{0.400pt}}
\put(731.0,639.0){\rule[-0.200pt]{4.818pt}{0.400pt}}
\put(915.0,660.0){\rule[-0.200pt]{0.400pt}{2.891pt}}
\put(905.0,660.0){\rule[-0.200pt]{4.818pt}{0.400pt}}
\put(905.0,672.0){\rule[-0.200pt]{4.818pt}{0.400pt}}
\put(1089.0,686.0){\rule[-0.200pt]{0.400pt}{2.650pt}}
\put(1079.0,686.0){\rule[-0.200pt]{4.818pt}{0.400pt}}
\put(1079.0,697.0){\rule[-0.200pt]{4.818pt}{0.400pt}}
\put(1262.0,697.0){\rule[-0.200pt]{0.400pt}{3.373pt}}
\put(1252.0,697.0){\rule[-0.200pt]{4.818pt}{0.400pt}}
\put(1252.0,711.0){\rule[-0.200pt]{4.818pt}{0.400pt}}
\put(1436.0,706.0){\rule[-0.200pt]{0.400pt}{2.650pt}}
\put(1426.0,706.0){\rule[-0.200pt]{4.818pt}{0.400pt}}
\put(1426.0,717.0){\rule[-0.200pt]{4.818pt}{0.400pt}}
\sbox{\plotpoint}{\rule[-0.500pt]{1.000pt}{1.000pt}}%
\put(1262,207){\makebox(0,0)[r]{New Bound}}
\multiput(1284,207)(20.756,0.000){4}{\usebox{\plotpoint}}
\put(1350,207){\usebox{\plotpoint}}
\put(220,113){\usebox{\plotpoint}}
\multiput(220,113)(7.474,19.363){3}{\usebox{\plotpoint}}
\multiput(242,170)(7.903,19.192){3}{\usebox{\plotpoint}}
\multiput(263,221)(9.541,18.432){5}{\usebox{\plotpoint}}
\multiput(307,306)(11.986,16.945){7}{\usebox{\plotpoint}}
\multiput(394,429)(16.317,12.827){11}{\usebox{\plotpoint}}
\multiput(567,565)(19.369,7.458){9}{\usebox{\plotpoint}}
\multiput(741,632)(20.325,4.205){8}{\usebox{\plotpoint}}
\multiput(915,668)(20.606,2.487){9}{\usebox{\plotpoint}}
\multiput(1089,689)(20.697,1.555){8}{\usebox{\plotpoint}}
\multiput(1262,702)(20.739,0.834){8}{\usebox{\plotpoint}}
\put(1436,709){\usebox{\plotpoint}}
\put(1262,162){\makebox(0,0)[r]{Old Bound}}
\multiput(1284,162)(41.511,0.000){2}{\usebox{\plotpoint}}
\put(1350,162){\usebox{\plotpoint}}
\put(220,113){\usebox{\plotpoint}}
\multiput(220,113)(15.662,38.443){2}{\usebox{\plotpoint}}
\put(252.08,189.56){\usebox{\plotpoint}}
\multiput(263,214)(20.198,36.265){2}{\usebox{\plotpoint}}
\multiput(307,293)(25.751,32.559){4}{\usebox{\plotpoint}}
\multiput(394,403)(35.395,21.687){5}{\usebox{\plotpoint}}
\multiput(567,509)(40.456,9.300){4}{\usebox{\plotpoint}}
\multiput(741,549)(41.377,3.329){4}{\usebox{\plotpoint}}
\multiput(915,563)(41.494,1.192){4}{\usebox{\plotpoint}}
\multiput(1089,568)(41.508,0.480){4}{\usebox{\plotpoint}}
\multiput(1262,570)(41.510,0.239){5}{\usebox{\plotpoint}}
\put(1436,571){\usebox{\plotpoint}}
\end{picture}
\caption{Simulation results and theoretical predictions for 10000 bins}
\label{fig_simulations}
\end{figure}
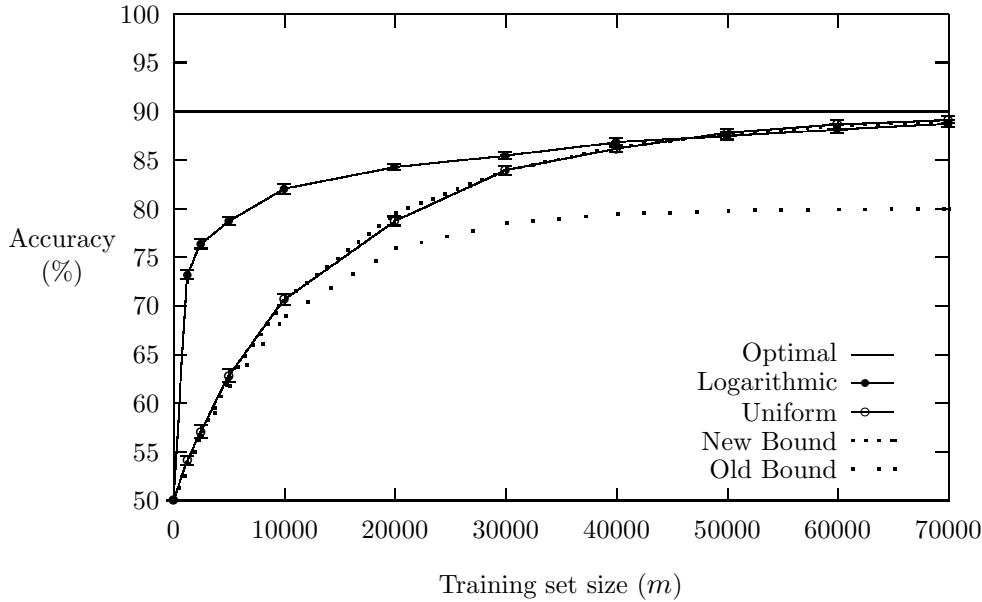

\subsection{Results}

Figure~\ref{fig_simulations} shows five traces of accuracy
as the volume of training data is varied.
The lowest curve shows the old bound which can be achieved
using the results in Lauer (1995), as represented by
equation~(\ref{eqn_old_bound}).  The other dotted curve shows
the expected accuracy predicted using equation~(\ref{eqn_ea_bound})
as approximated by the program described in section~\ref{sec_computable}.
The two further curves (with confidence interval bars) then show
the results of simulations, using uniform and logarithmic bin
distributions.

As can be seen, the new bound given in this paper is accurate
for uniform bin probabilities.  However, when the bins are
logarithmically distributed learning converges significantly
more quickly, as suggested by the reasoning about expected
number of relevant training instances (see section~\ref{sec_skewed}).
Perhaps surprisingly though, the logarithmic distribution
appears to eventually fall behind the uniform one once there is plenty
of data.  This might be explained by the presence of very rare bins
in the logarithmic distribution which thus take longer to learn.
Both these observations are crucial to reasoning
about data requirements for {\sc sll}.

\section{Conclusion}

If commercial {\sc nlp} systems are to be developed from the current
batch of research prototypes for {\sc sll}, then a predictive theory
of the data requirements of such systems is necessary.  In this paper
I have explored the dependence of the expected accuracy of a simple
statistical learner on the volume of training data.  When the
probability distribution of inputs is uniform, I have shown how to
compute the expected accuracy, a result backed up by simulations.
In particular, an average of four training instances per bin can
be expected to yield an error rate only 50\% worse than the
optimal error rate.

When the distribution is non-uniform, simulations show that
convergence can be much more rapid.  Error rates only 50\% worse
than optimal result from only three training instances per
bin.  However,
when data is abundant,
non-uniform distributions result in higher error rates
than the estimate produced by assuming uniformity.

\section{Acknowledgements}

I am grateful to Mark Johnson, without whom this work
would not exist, and also to Robert Dale, Mark Dras, Mike Johnson and
John Potter.  Financial support is gratefully acknowledged from the
Australian Government and the Microsoft Institute.


\begin{thebibliography}{}

\bibitem{} Brent, Michael.
\newblock 1993.
\newblock From Grammar to Lexicon: Unsupervised Learning of Lexical Syntax.
\newblock In {\em Computational Linguistics, {\bf Vol 19(2)}},
pp243-62.

\bibitem{} Charniak, Eugene.
\newblock 1993.
\newblock {\em Statistical Language Learning}.
\newblock MIT Press, Cambridge, MA.

\bibitem{} Dagan,~I. and Itai,~A.
\newblock 1990.
\newblock A Statistical Filter For Resolving Pronoun References.
\newblock In {\em Proceedings of the Seventh Israeli Conference on
Artificial Intelligence and Computer Vision}, Ramat Gan, Israel. pp125-35.

\bibitem{} de~Haan, Pieter.
\newblock 1992.
\newblock Optimum Corpus Sample Size?
\newblock In Leitner, Gerhard (ed.) {\em New Directions in
English Language Corpora}.
\newblock  Mouton de Gruyter, Berlin.

\bibitem{} Hindle,~D. and Rooth,~M.
\newblock 1993.
\newblock Structural Ambiguity and Lexical Relations.
\newblock In {\em Computational Linguistics {\bf Vol. 19(1)}},
pp103-20.

\bibitem{} Lauer,~Mark.
\newblock 1995.
\newblock How Much Is Enough? Data Requirements for Statistical NLP.
\newblock In {\em Proceedings of the 2nd Conference of the
Pacific Association for Computational Linguistics}, Brisbane, Australia.
cmp-lg/9509001

\bibitem{} Lauer,~M. and Dras,~M.
\newblock 1994.
\newblock A Probabilistic Model of Compound Nouns.
\newblock In {\em Proceedings of the 7th Australian Joint
Conference on Artificial Intelligence}, Armidale, NSW, Australia.
World Scientific Press, pp474-81. cmp-lg/9409003

\bibitem{} Marcus,~M., Marcinkiewicz,~M.~A. and Santorini,~B.
\newblock 1993.
\newblock Building a Large Annotated Corpus of English: The Penn Treebank.
\newblock In {\em Computational Linguistics {\bf Vol 19(2)}},
pp313-30.

\bibitem{} Yarowsky, David.
\newblock 1992.
\newblock Word-Sense Disambiguation Using Statistical Models of
Roget's Categories Trained on Large Corpora.
\newblock In {\em Proceedings of COLING-92}, France, pp454-60.

\end{thebibliography}
\end{document}